\documentclass[runningheads]{llncs}

\usepackage{graphicx}
\usepackage{epstopdf}
\usepackage{amsmath}
\usepackage{amsfonts}
\usepackage{cite}
\usepackage{hyperref}
\newcommand{\rpm}{\raisebox{.2ex}{$\scriptstyle\pm$}}

\hypersetup{colorlinks,urlcolor=blue}

\newcommand{\mycomment}[1]{}

\begin{document}
\title{Harnessing Uncertainty in Domain Adaptation for MRI Prostate Lesion Segmentation}
\titlerunning{Harnessing Uncertainty in Domain Adaptation}
%
\author{Eleni Chiou \inst{1, 2}
\and Francesco Giganti \inst{3, 4} 
\and Shonit Punwani  \inst{5} 
\and Iasonas Kokkinos \inst{2}
\and Eleftheria Panagiotaki \inst{1, 2}}


\authorrunning{E. Chiou et al.}
%
\institute{Centre for Medical Image Computing, UCL, London, UK
\and Department of Computer Science, UCL, London, UK 
\and Department of Radiology, UCLH NHS Foundation Trust, London, UK 
\and Division of Surgery \& Interventional Science, UCL, London, UK
\and Centre for Medical Imaging, Division of Medicine, UCL, London, UK
\email{eleni.chiou.17@ucl.ac.uk}}
\maketitle              
\begin{abstract}
The need for training data can impede the adoption of novel imaging modalities for learning-based medical image analysis. Domain adaptation methods partially mitigate this problem by translating training data from a related source domain to a novel target domain, but typically assume that a one-to-one translation is possible. Our work addresses the challenge of adapting to {\emph{a more informative target domain}} where multiple target samples can emerge from a single source sample. In particular we consider translating from mp-MRI to VERDICT, a richer MRI modality involving an optimized acquisition protocol for cancer characterization. We explicitly account for the inherent uncertainty of this mapping and exploit it to generate multiple outputs conditioned on a single input. Our results show that this  allows us to extract systematically better image representations for the target domain,  when used in tandem with both simple, CycleGAN-based baselines, as well as more powerful approaches that integrate discriminative segmentation losses  and/or residual adapters. When compared to its deterministic counterparts, our approach yields substantial improvements across a broad range of dataset sizes, increasingly strong baselines, and evaluation measures.

\keywords{Domain adaptation, Image synthesis, GANs, Segmentation, MRI}
\end{abstract}

\section{Introduction}
Domain adaptation can be used to exploit training samples from an existing, densely-annotated domain within a novel, sparsely-annotated domain, by bridging the  differences between the two domains. 
This can facilitate the training of powerful convolutional neural networks (CNNs) for novel medical imaging modalities or acquisition protocols, effectively compensating for the limited amount of training data available to train CNNs in the new domain.

The  assumption underlying most domain adaptation methods is that one can align the two domains either by extracting domain-invariant representations (features), or by establishing a `translation' between the two domains at the signal level, where in any domain the `resident' and the translated signals are statistically indistinguishable.

In particular for medical imaging,~\cite{Ren_MICCAI_18} and~\cite{Kamnitsas_IPMI_17} rely on adversarial training to align the feature distributions between the source and the target domain for medical image classification and segmentation respectively. Pixel-level distribution alignment is performed by~\cite{Jiang_MICCAI_18, Zhang_MICCAI_18, Cai_MedIA_19, Zhang_CVPR_18}, who use CycleGAN~\cite{CycleGAN} to map source domain images to the style of the target domain; they further combine the translation network with a task-specific loss to penalize semantic inconsistency between the source and the synthesized images. The synthesized images are used to train models for image segmentation in the target domain. Ouyang et al.~\cite{Quyang_MICCAI_19} perform adversarial training to learn a shared, domain-invariant latent space which is exploited during segmentation. They show that their approach is effective in cases where target-domain data is scarce. Similarly,~\cite{Yang_MICCAI_19} embed the input images from both domains onto a domain-specific style space and a shared content space. Then, they use the content-only images to train a segmentation model that operates well in both domains. However, their approach does not necessarily preserve crucial semantic information in the content-only images.

These methods rely on the strong assumption that the two domains can be aligned - our work shows that accuracy gains can be obtained by acknowledging that this can often be only partially true, and mitigating the resulting challenges. As a natural image example, an image taken at night can have many day-time counterparts, revealed by light; similarly in medical imaging, a better imaging protocol can reveal structures that had previously passed unnoticed. In technical terms, the translation can be one-to-many, or, stated in  probabilistic terms, multi-modal~\cite{Zhu_NIPS_17, Huang_ECCV_18, Lee_ECCV_18}. Using a one-to-one translation network in such a setting can harm performance, since the translation may predict the mean of the underlying  multi-modal distribution, rather than provide diverse, realistic samples from it. 

\begin{figure}[!t]
\includegraphics[width=\textwidth]{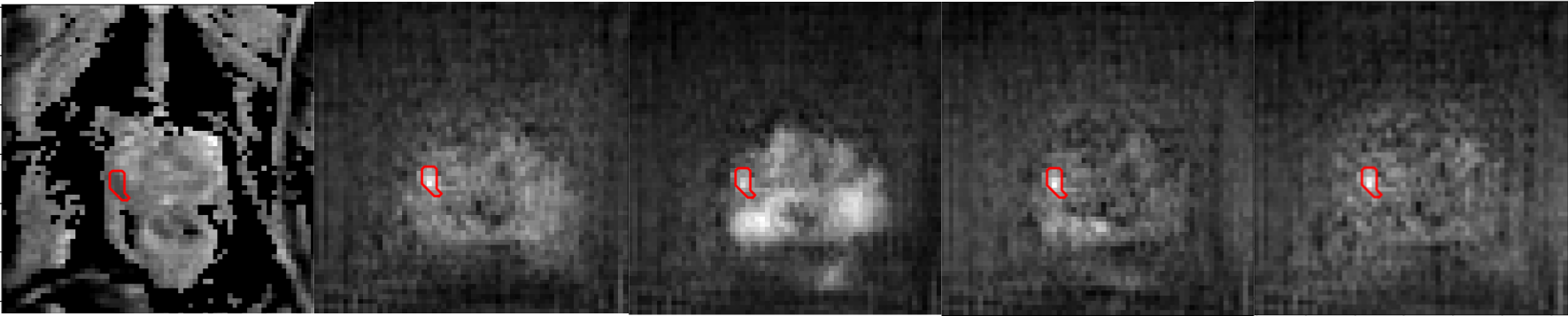}
\caption{One-to-many mapping from one mp-MRI image (left) to four VERDICT-MRI translations: our network can generate samples with both local and global structure variation, while at the same time preserving the critical structure corresponding to the prostate lesion, shown as a red circle. We note that the lesion area is annotated by a physician on the leftmost image, but is not used as input to the translation network - instead the translation network learns to preserve lesion structures thanks to the end-to-end discriminative training (details in text).}
\label{transl_many_samples}
\end{figure}

In our work we accommodate the inherent uncertainty in the cross-domain mapping and, as shown in Figure \ref{transl_many_samples}, generate multiple outputs conditioned on a single input, thereby allowing for better generalization of the segmentation network in the target domain.
As in recent studies~\cite{Jiang_MICCAI_18, Zhang_MICCAI_18, Cai_MedIA_19, Zhang_CVPR_18}, we use GANs~\cite{Goodfellow_NIPS_14} to align the source and target domains, but go beyond their one-to-one, deterministic mapping approaches. In addition, inspired by~\cite{Jiang_MICCAI_18, Zhang_MICCAI_18, Cai_MedIA_19, Hoffman_ICML_18}, we enforce semantic consistency between the real and synthesized images by exploiting source-domain lesion segmentation supervision to train target-domain networks operating on the synthesized images. This results in training networks that can generate diverse outputs while at the same time preserving critical structures - such as the lesion area in Figure~\ref{transl_many_samples}. We further accommodate the statistical discrepancies between real and synthesized data by introducing residual adapters (RAs)~\cite{Rebuffi_CVPR_18, Chiou_ISMRM_20} in the segmentation network. These capture domain-specific properties and allow the segmentation network to generalize better across the two domains.

We demonstrate the effectiveness of our approach in prostate lesion segmentation and an advanced diffusion weighted (DW)-MRI method called VERDICT-MRI (Vascular, Extracellular and Restricted Diffusion for Cytometry in Tumors). VERDICT-MRI is a non-invasive imaging technique for cancer microstructure characterisation~\cite{Verdict1, Verdict2, Verdict3}. The method has been recently in clinical trial to supplement standard multi-parametric (mp)-MRI for prostate cancer diagnosis. Compared to the naive DW-MRI from mp-MRI acquisitions, VERDICT-MRI has a richer acquisition protocol to probe the underlying microstructure and reveal changes in tissue features similar to histology. A recent study~\cite{Verdict3} has demonstrated that the intracellular volume fraction (FIC) maps obtained with VERDICT-MRI differentiate better benign and malignant lesions compared to the apparent diffusion coefficient (ADC) map obtained with the naive DW-MRI from mp-MRI acquisitions. However, the limited amount of available labeled training data does not allow the training of robust deep neural networks that could directly exploit the information in the raw VERDICT-MRI~\cite{Chiou_MLMI_18}. On the other hand, large scale clinical mp-MRI datasets exist~\cite{prostateX, promis}. As shown experimentally, our approach largely improves the generalization capabilities of a lesion segmentation model on VERDICT-MRI by exploiting labeled mp-MRI data.

\section{Method}

\begin{figure}[!t]
\includegraphics[width=\textwidth]{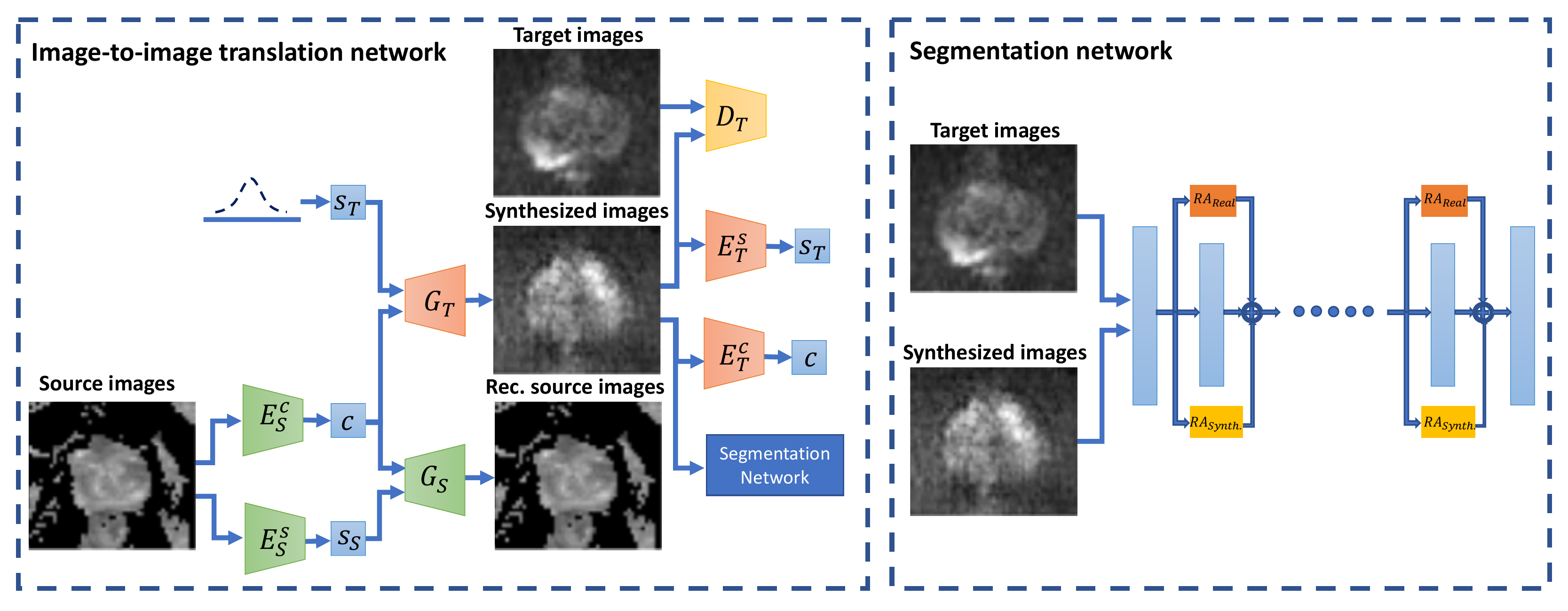}
\caption{Overview of our domain adaptation framework: we train a noise-driven domain translation network in tandem with a discriminatively supervised segmentation network in the target domain; GAN-type losses align the translated samples with the target distribution, while residual adapters allow the segmentation network to compensate for remaining discrepancies. Please see text for details.}
\label{block}
\end{figure}

Our approach relies on a unified network for cross-modal image synthesis and segmentation, that is trained end-to-end with a combination of objective functions. As shown in Figure~\ref{block}, at the core of this network is an image-to-image translation network that maps images from the source ('S') to the target ('T') domain. The translation network is trained in tandem with a segmentation network that operates in the target domain, and is trained with both the synthesized and the few real annotated target-domain images.
Beyond these standard components, our approach relies on three additional components: firstly, we sample a latent variable from a Gaussian distribution when translating to the target domain; this represents structures that cannot be accounted by a deterministic mapping, and can result in one-to-many translation when needed. Secondly, we introduce residual adapters (RAs) to a common backbone network for semantic segmentation, allowing the discriminative training to accommodate any remaining discrepancies between the real and synthesized target domain images. 
Finally, we use a dual translation network from the target to the source domain, allowing us to use cycle-consistency in domain adaptation~\cite{Liu_NIPS_17, CycleGAN, Huang_ECCV_18}; the cycle constraint allows us to disentangle the deterministic, transferable part from the stochastic, non-transferable part, which is filled in by Gaussian sampling, as mentioned earlier.

\subsection{Problem formulation}
Having provided a broad outline of our method, we now turn to a more detailed technical description. 
We consider the problem of domain adaptation in prostate lesion segmentation. We assume that the source domain, $\mathcal{X}_S$, contains $N_S$ images, ${x}_S \in \mathcal{X}_S$, with associated segmentation masks, ${y}_S \in \mathcal Y_S$. Similarly, the sparsely labeled target domain, $\mathcal{X}_T$, consists of $N_T$ images, $x_T \in \mathcal{X}_T$. A subset $\tilde{\mathcal{X}}_T$ of $\mathcal{X}_T$ comes with associated segmentation masks, $y_T \in \mathcal Y_T$.
The proposed framework consists of two main components, i.e. an image-to-image translation network and a segmentation network described below.

\subsection*{Segmentation Network}
The segmentation network (Fig.~\ref{block}), $Seg$, operates on image-label pairs of both real, $\mathcal{X}_T$, and synthesized data, $\mathcal{X}_{S\rightarrow T}$, translated from source to target. An encoder-decoder network~\cite{Chen_ECCV_18, Unet} is the main backbone which serves both domains. To compensate further for differences in the feature statistics of real and synthesized data we install residual adapter modules~\cite{Rebuffi_CVPR_18} in parallel to each of the convolutional layer of the backbone. Introducing residual adapters ensures that most of the parameters stay the same with the network, but also that the new unit introduces a small, but effective modification that accommodates the remaining statistical discrepancies of the two domains.

More formally, let $\phi_l$ be a convolutional layer in the segmentation network and $\mathbf F^l\in \mathbb R^{k \times k \times C_i \times C_o} $ be a set of filters for that layer, where $k\times k$ is the kernel size and $C_i$, $C_o$ are the number of input and output feature channels respectively. Let also $\mathbf Z_i^l \in \mathbb R^{1 \times 1 \times C_i \times C_o} $ be a set of domain-specific residual adapter filters of domain $i$, where $i \in \{1, 2\}$, installed in parallel with the existing set of filters $\mathbf F_l$. Given an input tensor $\mathbf x_l \in \mathbb R^{H \times W \times C_i} $, the output $\mathbf y_l \in \mathbb R^{H \times W \times C_o} $ of layer $l$ is given by 

\begin{equation}
\mathbf{y}_l= \mathbf{F}^l * \mathbf{x} + \mathbf{Z}_i^l * \mathbf{x}.
\end{equation}

\noindent We train the segmentation network by optimizing the following objective
\begin{equation}
\begin{aligned}
& \mathcal L_{Seg}(Seg, \tilde{\mathcal{X}_T}, \mathcal{Y}_T, \mathcal{X}_{S \rightarrow T}, \mathcal{Y}_S) = \\ & \quad \quad \mathcal L_{DSC}(Seg, \tilde{\mathcal{X}_T}, \mathcal{Y}_T) + 
\mathcal L_{DSC}(Seg, \mathcal{X}_{S \rightarrow T}, \mathcal{Y}_S).
\end{aligned}
\end{equation}

\noindent The dice loss, $\mathcal L_{DSC}$, is given by 
\begin{equation}
\mathcal L_{DSC}(Seg,\mathcal{X},\mathcal{Y}) = - \frac {2 \sum_{(\mathbf{x}, \mathbf{y}) \in (\mathcal{X}, \mathcal{Y})} \sum_{k=1}^K Seg(\mathbf{x})_k \mathbf{y}_k}{\sum_{(\mathbf{x}, \mathbf{y}) \in (\mathcal{X}, \mathcal{Y})} \sum_{k=1}^K (Seg(\mathbf{x})_k^2 + \mathbf {y}_k^2)}, 
\end{equation}
where K the number of voxels in the input images. We adopt this objective function since it is a differentiable approximation of a criterion that is well-adapted to our task.

\subsection*{Diverse Image-to-Image Translation Network}
Recently, several studies~\cite{Zhu_NIPS_17, Huang_ECCV_18} have pointed out that cross-domain mapping is inherently multi-modal and proposed approaches to produce multiple outputs conditioned on a single input. Here we use MUNIT~\cite{Huang_ECCV_18} to illustrate the key idea. As it is illustrated in Figure \ref{block} the image-to-image translation network consists of content encoders $E_S^c$, $E_T^c$, style encoders $E_S^s$, $E_T^s$, generators $G_S$, $G_T$ and domain discriminators $D_S$, $D_T$ for both domains. The content encoders $E_S^c$, $E_T^c$ map images from the two domains onto a domain-invariant content space $\mathcal{C}$ ($E_S^c: \mathcal{X}_S \rightarrow \mathcal{C}$, $E_T^c: \mathcal{X}_T \rightarrow \mathcal{C}$) and the style encoders $E_S^s$, $E_T^s$ map the images onto domain-specific style spaces $\mathcal{S}_S$ ($E_S^s: \mathcal{X}_S \rightarrow \mathcal{S}_S$) and $\mathcal{S}_T $ ($E_T^s: \mathcal{X}_T \rightarrow \mathcal{S}_T$). The content code can be understood as the underlying anatomy which is the information that we want transfer during the translation while the style codes capture information related to the imaging modalities. Image-to-image translation is performed by combining the content code extracted from a given input and a random style code sampled from the target-style space. For instance, to translate an image $x_S \in \mathcal{X}_S$ to $\mathcal{X}_T$ we first extract its content code $c = E_S^c(x_S)$. The generator $G_T$ uses the extracted content code $c$ and a randomly drawn style code $s_T \in \mathcal{S}_T$ to produce the final output $x_{S \rightarrow T} = G_T(c, s_T)$. By sampling random style codes from the style spaces $\mathcal{S}_S$ and $\mathcal{S}_T$ the generators ${G}_S$ and ${G}_T$ are able to produce diverse outputs.
We train the networks with a loss function that consists of domain adversarial, self-reconstruction, latent reconstruction, cycle-consistency  and segmentation losses.

\noindent \textbf{Domain adversarial loss}. We utilize GANs to match the distribution between the synthesized and the real images of the two domains. The adversarial discriminators $D_T$, $D_S$ aim at discriminating between real and synthesized images, while the generators $G_T$, $G_S$ aim at generating realistic images that fool the discriminators. For $G_T$ and $D_T$ the GAN loss is defined as
\noindent \begin{equation}
\begin{aligned}
&\mathcal L_{GAN}^T(E_S^c, G_T, D_T, \mathcal{S}_T, \mathcal{X}_{S}) = \\
& \quad \mathbb{E}_{ x_S \sim \mathcal X_S, s_T \sim \mathcal S_T}[\log(1-D_T(G_T(E_S^c(x_S), s_T)))] 
+ \mathbb{E}_{x_T \sim \mathcal X_T}[\log(D_T(x_T))].
\end{aligned}
\end{equation} 

\noindent \textbf{Self-reconstruction loss}.
Given the encoded content and style codes of a source-domain image the generator ${G}_S$ should be able to decode them back to the original one.
\begin{equation}
\begin{aligned}
\mathcal L_{recon}^S(G_S, E_S^s, E_S^c, \mathcal{X}_S) = \mathbb{E}_{x_S \sim \mathcal X_S}[\|G_S(E_S^c(x_S), E_S^s(x_S))-x_S\|_1].
\end{aligned}
\end{equation} 
\noindent \textbf{Latent reconstruction loss}. 
To encourage the translated image to preserve the content of the source image, we require that a latent code $c$ sampled from the latent distribution can be reconstructed after decoding and encoding.
\begin{equation}
\begin{aligned}
& \mathcal L_{recon}^{c_S}(E_S^c, G_T, E_T^c, \mathcal{X}_S, S_T) = \\ 
& \quad \quad \mathbb{E}_{x_S \sim \mathcal X_S, s_T \sim \mathcal S_T}[\|E_T^c(G_T(E_S^c(x_s), s_T))- E_S^c(x_s)\|_1].
\end{aligned}
\end{equation} 

\noindent Similarly, to align the style representation with a Gaussian prior distribution, we enforce the same constrain for the latent style code.
\begin{equation}
\begin{aligned}
& \mathcal L_{recon}^{s_T}(E_S^c, G_T, E_T^s, \mathcal{X}_S, S_T) = 
\\ & \quad \quad \mathbb{E}_{x_S \sim \mathcal X_S, s_T \sim \mathcal S_T} [\|E_T^s(G_T(E_S^c(x_s), s_T))- s_T)\|_1].
\end{aligned}
\end{equation} 

\noindent \textbf{Cycle-consistency loss}. 
To facilitate training we enforce cross-cycle consistency which implies that if we translate an image to the target domain and then translate it back to the source domain using the extracted source-domain style code, we should be able to obtain the original image. 
\begin{equation}
\begin{aligned}
& \mathcal L_{cyc}^{S}(E_S^c, E_S^s, G_T, E_T^c, G_S, \mathcal{X}_S, S_T) = 
\\ & \quad \quad \mathbb{E}_{x_S \sim \mathcal X_S, s_T \sim \mathcal S_T} [\|G_S(E_T^c(G_T(E_S^c(x_S), s_T)), E_S^s(x_S))-x_S\|_1].
\end{aligned}
\end{equation} 
\noindent $\mathcal L_{GAN}^S$, $\mathcal L_{recon}^T$, $\mathcal L_{recon}^{c_T}$, $\mathcal L_{recon}^{s_S}$, $\mathcal L_{cyc}^{T}$ are defined in a similar way. 

\noindent \textbf{Segmentation loss}. To enforce the generator to preserve the lesions, we enrich the network with segmentation supervision on the synthesized images. The segmentation loss on the synthesized images is given by
\begin{equation}
\mathcal L_{Seg}^{Synth}(Seg, G_T, E_S^c, \mathcal{X}_S, \mathcal{Y}_S, \mathcal S_T) = \mathcal L_{DSC}(Seg, G_T(E_S^c(\mathcal X_S), S_T), \mathcal{Y}_S).
\end{equation}

\noindent The full objective is given by
\begin{equation} 
\begin{aligned}
& \min_{\substack{E_S^c, E_S^s, E_T^c, E_T^s, G_S, G_T}} \max_{ D_S, D_T} 
 \lambda_{GAN} (\mathcal L_{GAN}^S + \mathcal L_{GAN}^T) + \lambda_{x} (\mathcal L_{recon}^S + \mathcal L_{recon}^T) \\
&\quad \quad \quad +\lambda_{c} (\mathcal L_{recon}^{c_S} + \mathcal L_{recon}^{c_T})
+\lambda_{s} (\mathcal L_{recon}^{s_S} + \mathcal L_{recon}^{s_T}) \\
&\quad \quad \quad \quad \quad +\lambda_{cyc} (\mathcal L_{cyc}^{S} + \mathcal L_{cyc}^{T})
+\mathcal L_{Seg}^{Synth}, 
\end{aligned}
\end{equation}
where $\lambda_{GAN}$, $\lambda_{x}$, $\lambda_{c}$, $\lambda_{s}$, $\lambda_{cyc}$ are weights that control the importance of each term.

\subsection{Implementation details}
We implement our model using Pytorch~\cite{pytorch}. The content encoders consist of several convolutional layers and residual blocks followed by instance normalization \cite{IN}. The style encoders consist of convolutional layers followed by fully connected layers. The decoders include residual blocks followed by upsampling and convolutional layers. The residual blocks are followed by adaptive instance normalization (AdaIN)~\cite{AdaIN} layers to adjust the style of the output image. The affine parameters of AdaIN are generated by a multilayer perceptron from a given style code. The discriminators consist of several convolutional layers. The encoder of the segmentation network is a standard ResNet~ \cite{resnet} consisting of several convolutional layers while the decoder consists of several upsampling and convolutional layers. For training we use Adam optimizer, a batch size of 32 and a learning rate of 0.0001. We make our code available at \url{https://github.com/elchiou/DA}.

\subsection{Datasets}
\textbf{VERDICT-MRI}: We use VERDICT-MRI data collected from $60$ men with a suspicion of cancer. VERDICT-MRI images were acquired with pulsed-gradient spin-echo sequence (PGSE) using an optimized imaging protocol for VERDICT prostate characterization with 5 b-values ($90$, $500$, $1500$, $2000$, $3000 \ \rm{s/mm^2}$) in 3 orthogonal directions ~\cite{acq_prot}. Images with $\rm{b} = 0 \ \rm{s/mm^2}$ were also acquired before each b-value acquisition. The DW-MRI sequence was acquired with a voxel size of $1.25 \times 1.25 \times 5 \ \rm{mm^3}$, $5 \ \rm{mm}$ slice thickness, $14$ slices, and field of view of $220 \times 220 \ \rm{mm^2}$ and the images were reconstructed to a $176 \times 176$ matrix size. A dedicated radiologist, highly experienced in prostate mp-MRI, contoured the lesions on VERDICT-MRI using mp-MRI for guidance. 

\noindent\textbf{DW-MRI from mp-MRI acquisition}: We use DW-MRI data from the ProstateX challenge dataset~\cite{prostateX} which consists of training mp-MRI data acquired from $204$ patients. The DW-MRI data were acquired with a single-shot echo planar imaging sequence with a voxel size of $2 \times 2 \times 3.6\ \rm{mm^3}$, $3.6\  \rm{mm}$ slice thickness. Three b-values were acquired ($50, 400, 800  \  \rm{s/mm^2}$ ), and subsequently, the ADC map and a b-value image at $\rm{b} = 1400\  \rm{s/mm^2}$ were calculated by the scanner software. In this study, we use DW-MRI data from $80$ patients. Since the ProstateX dataset provides only the position of the lesion, a dedicated radiologist manually annotated the lesions on the ADC map using as reference the provided position of the lesion.

\begin{table}[t]
\center
\caption{Average recall, precision, dice similarity coefficient (DSC), and average precision (AP) across $5$ folds. The results are given in mean (\rpm std) format.}\label{table}
\begin{tabular}{|l|l|l|l|l|}
\hline
 Model                                   & Recall       & Precision       & DSC          & AP            \\ \hline
VERDICT-MRI only                         & 67.1 (\rpm 14.2)    & 59.6 (\rpm 11.5)   & 62.4 (\rpm 13.4)  &  63.5 (\rpm 13.1)  \\ \hline
Finetuning                               & 68.4 (\rpm 12.4)    & 62.5 (\rpm 13.5)   &  64.7 (\rpm 11.2) &  65.8 (\rpm 14.7)  \\ \hline
RAs                                      & 66.6 (\rpm 11.6)    & 67.0 (\rpm 8.8)    &  65.7 (\rpm 10.2) &  66.6 (\rpm 12.6)  \\ \hline
MUNIT                                    & 65.2 (\rpm 10.2)    & 64.2 (\rpm 13.7)   &  64.4 (\rpm 11.3) &  68.2 (\rpm 12.0) \\ \hline
CycleGAN + $\mathcal L_{Seg}^{Synth}$            & 64.5 (\rpm 10.4)    & 66.1 (\rpm 10.1)   &  64.8 (\rpm 8.7)  &  70.1 (\rpm 9.8)   \\ \hline
CycleGAN + $\mathcal L_{Seg}^{Synth}$ + RAs      & 60.9 (\rpm 10.7)    & \textbf{74.0} (\rpm 11.8)   &  66.6 (\rpm 13.6) &  71.6 (\rpm 11.3)  \\ \hline
MUNIT    + $\mathcal L_{Seg}^{Synth}$ (Ours)     & \textbf{71.8} (\rpm 7.8)     & 68.0 (\rpm 6.8)    &  69.8 (\rpm 7.9)  &  73.5 (\rpm 8.1)   \\ \hline
MUNIT + $\mathcal L_{Seg}^{Synth}$ + RAs (Ours)  & 69.2 (\rpm 8.6)     & 71.2 (\rpm 9.7)    &  \textbf{69.9} (\rpm 9.0)  &  \textbf{75.4} (\rpm 9.7)   \\ \hline
\end{tabular}
\end{table}

\begin{figure}[!h]
\centering
\begin{minipage}[c][1\width]{0.3\textwidth}
\centering
\includegraphics[width=\textwidth]{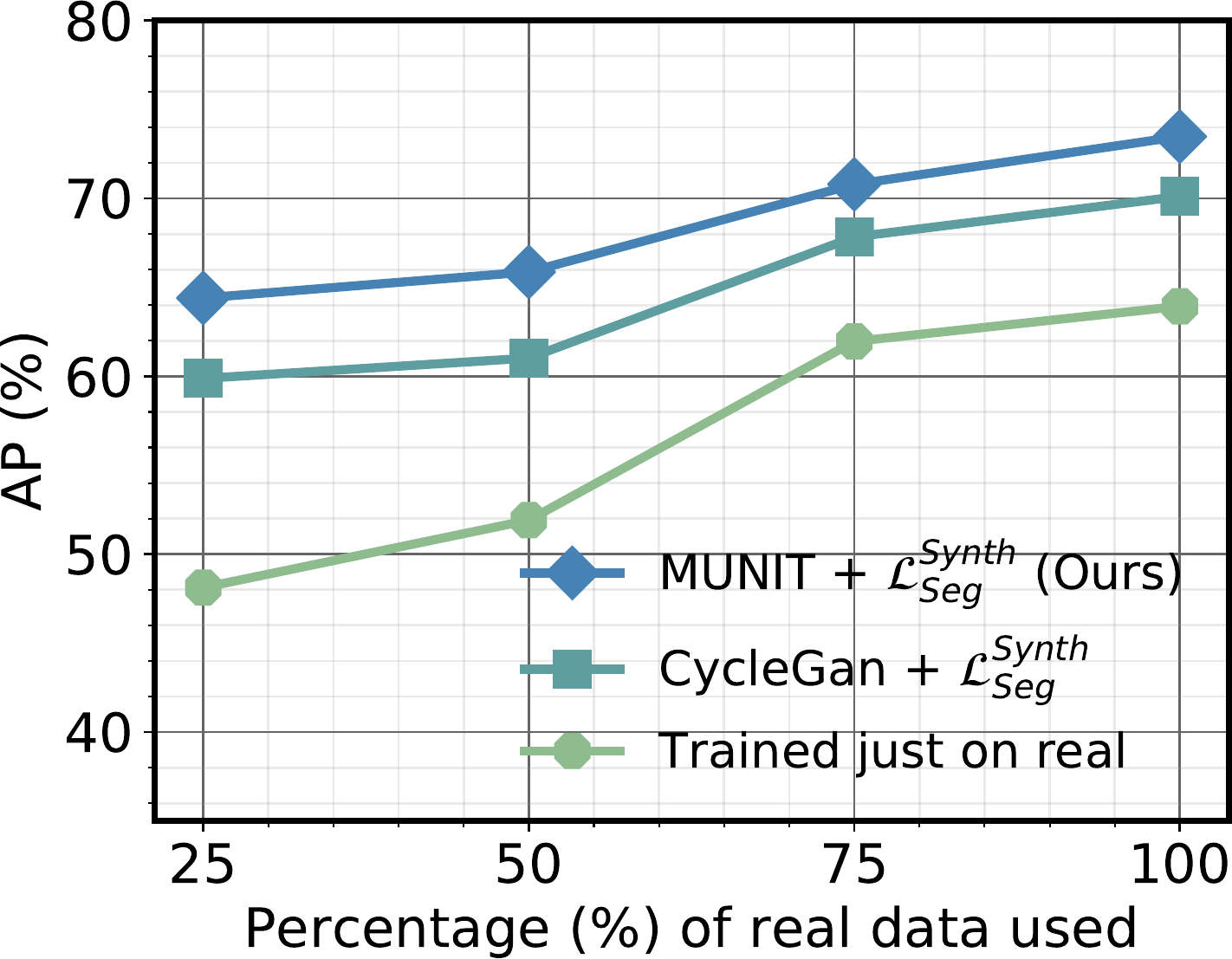}
\end{minipage}
\begin{minipage}[c][1\width]{0.3\textwidth}
\centering
\includegraphics[width=\textwidth]{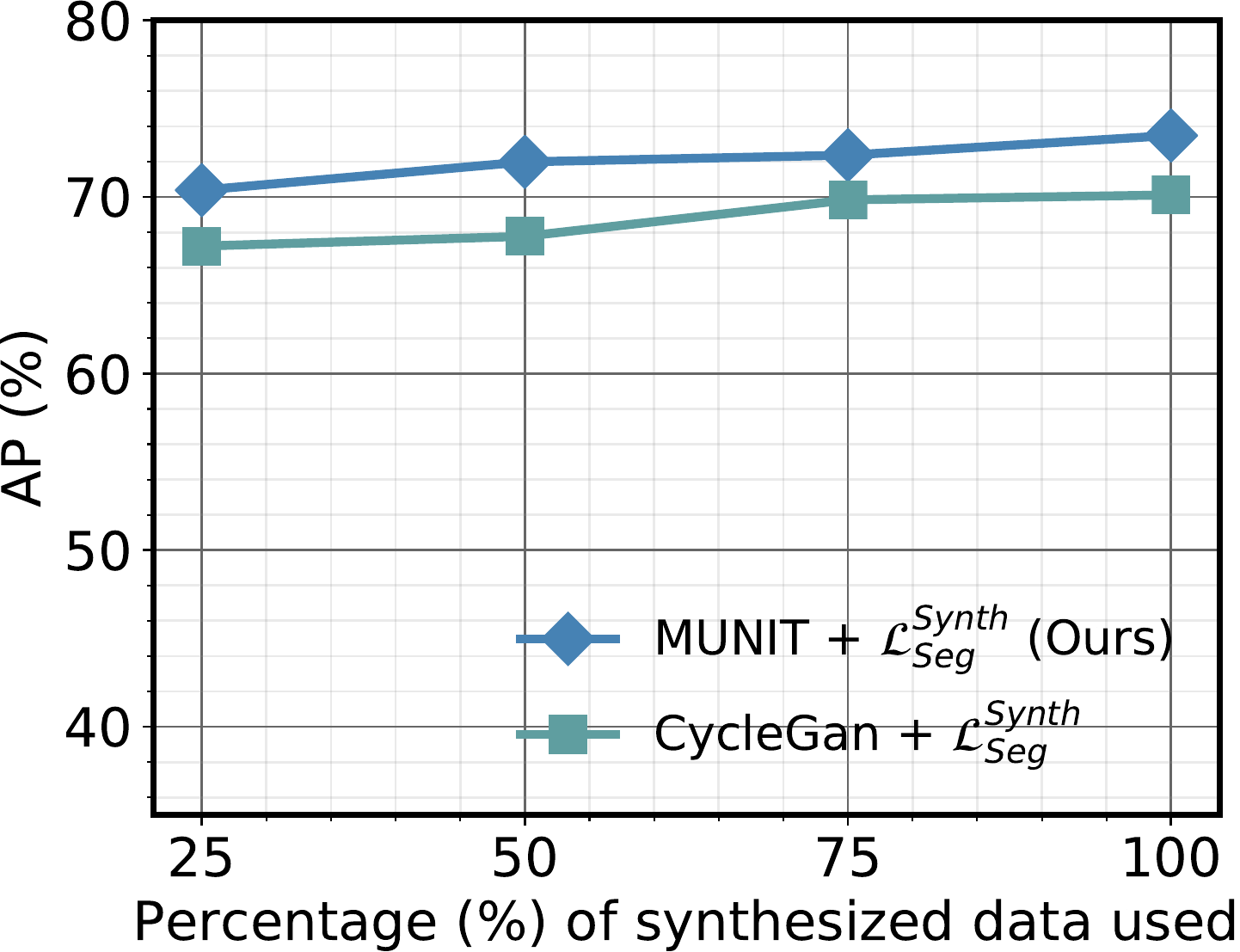}
\end{minipage}
\begin{minipage}[c][1\width]{0.3\textwidth}
\centering
\includegraphics[width=\textwidth]{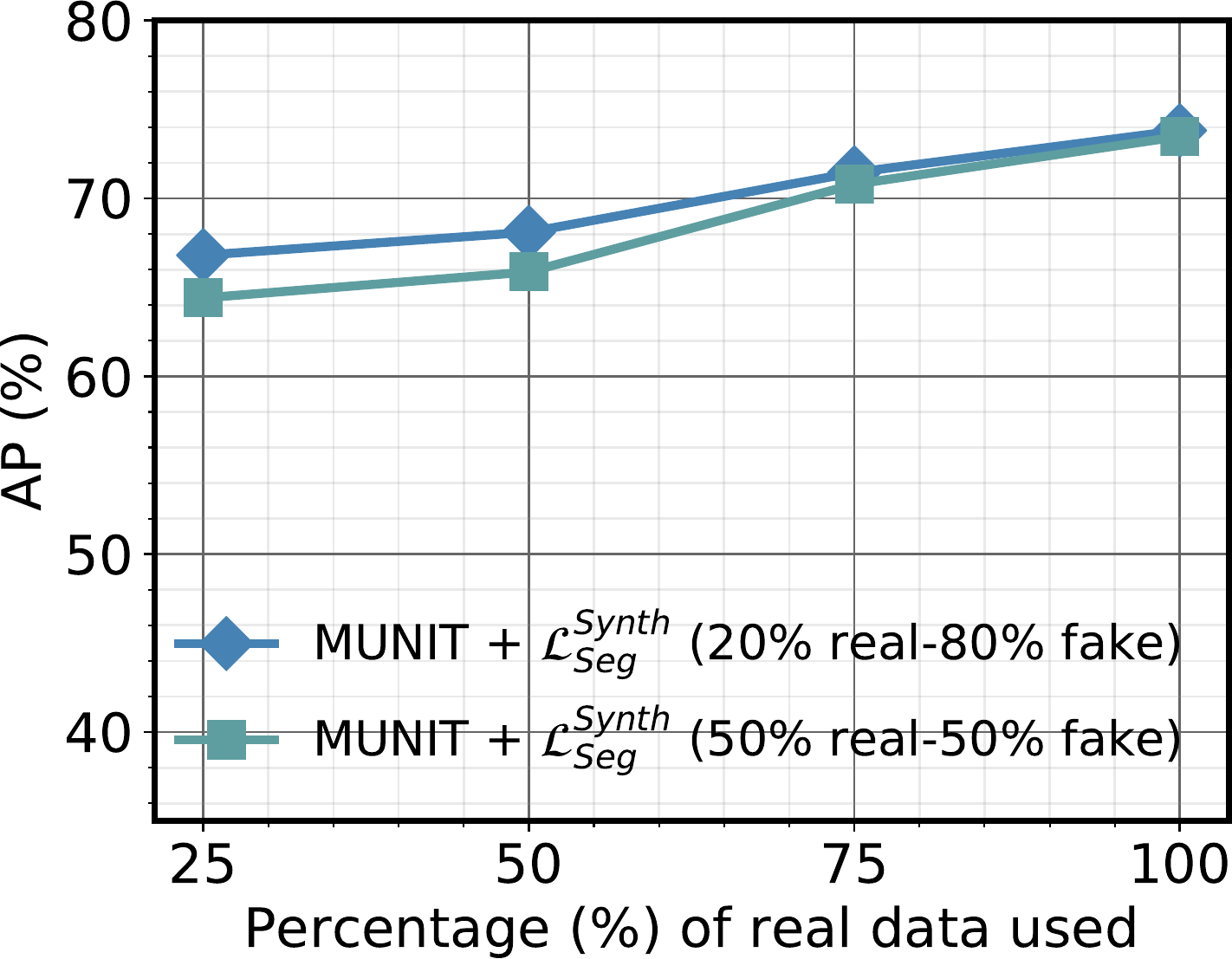}
\end{minipage}
\caption{Impact of the ratio of synthesized to real data on the performance. (Right) Average precision (AP) as a function of the percentage of real samples used given a constant number of synthesized ones. (Middle) AP as a function of the number of synthesized examples used given a constant number of real ones. (Left) AP as a function of the percentage of real data used given a constant number of synthesized ones. Here, the ratio of real to synthesized data in a mini-batch also varies during training.}\label{fig1}
\end{figure}

\section{Results}
In this section we evaluate the performance of our approach and the impact of the ratio of synthesized to real data on the performance. In the supplementary material we provide qualitative results and quantitative results related to the effect of sampling random style codes on the performance.

\noindent\textbf{Performance evaluation}. 
We first compare our approach to several baselines. i)VERDICT-MRI only: we train the segmentation network only on VERDICT-MRI. ii) Finetuning: we pre-train on mp-MRI and then perform finetuning using the VERDICT-MRI data. iii) RAs: we pre-train on mp-MRI, then we install RAs in parallel to each of the convolutional layers of the pre-trained network and update them using VERDICT-MRI. iv) MUNIT: we use MUNIT to map from source to target without segmentation supervision. v) CycleGAN + $\mathcal L_{Seg}^{Synth}$: we use CycleGAN and segmentation supervision to perform the translation, an approach similar to the one proposed in~\cite{Zhang_CVPR_18}. vi) CycleGAN + $\mathcal L_{Seg}^{Synth}$ + RAs: we use (v) for the translation and introduce RAs to the segmentation network. 
We evaluate the performance based on the average recall, precision, dice similarity coefficient (DSC), and average precision (AP). We report the results in Table \ref{table}. The proposed approach yields substantial improvements and outperforms all baselines including CycleGAN, which indicates that accommodating the uncertainty in the cross-domain mapping allows us to learn better representations for the target domain. Compared to the naive MUNIT without segmentation supervision, $\mathcal L_{Seg}^{Synth}$, our approach performs better since it successfully preserves the lesions during the translation. Finally, introducing RAs in the segmentation networks further improves the performance of both CyclgeGAN + $\mathcal L_{Seg}^{Synth}$ and MUNIT + $\mathcal L_{Seg}^{Synth}$. 

\noindent\textbf{Impact of the ratio of synthesized to real data on the performance.}
Using synthesized data is motivated by the fact that annotating large datasets can be challenging in medical applications. We therefore evaluate the impact of the ratio of synthesized to real data. To this end, we first vary the percentage of real data while keeping fixed the amount of synthesized data (Fig. \ref{fig1} (left)). We compare our approach to a segmentation network trained only on real data and to~\cite{Zhang_CVPR_18} where CycleGAN is used for the generation of synthesized data. Our approach outperforms both baselines. Figure \ref{fig1} (middle) shows the performance when we vary the percentage of synthesized samples while fixing the percentage of real ones. The AP of our approach increases as we increase the amount of synthesized data. The baseline also improves but we systematically outperform it. Figure \ref{fig1} (right) shows the performance of our approach when we vary the percentage of real data while fixing the percentage of synthesized. Here, we also vary the ratio of real to synthesized data in a mini-batch during training. Note that when the percentage of real data is small, a large ratio of synthesized to real data in the mini-batch delivers better results.

\section{Conclusion}
In this work we propose a domain adaptation approach for lesion segmentation. Our approach exploits the inherent uncertainty in the cross-domain mapping to generate multiple outputs conditioned on a single input allowing the extraction of richer representations for the task of interest in the target domain. We demonstrate the effectiveness of our approach in lesion segmentation on VERDICT-MRI, which is an advanced imaging modality for prostate cancer characterization. However, our approach is quite general can be applied in other application where the amount of labeled training data is limited.

\section*{Acknowledgments}
 This research is funded by EPSRC grand EP/N021967/1. We gratefully acknowledge the support of NVIDIA Corporation with the donation of the GPU used for this research.
\bibliographystyle{splncs04}
\bibliography{bibliography}

\end{document}